# Symbolic Representation and Classification of Logos

D S Guru[1] and N Vinay Kumar[1*]

[1]Department of Studies in Computer Science, University of Mysore, Mysore-570006
`dsg@compsci.uni-mysore.ac.in, vinaykumar.natraj@gmail.com`

**Abstract.** In this paper, a model for classification of logos based on symbolic representation of features is presented. The proposed model makes use of global features of logo images such as color, texture, and shape features for classification. The logo images are broadly classified into three different classes viz. logo image containing only text, an image with only symbol, and an image with both text and a symbol. In each class, the similar looking logo images are clustered using K-means clustering algorithm. The intra-cluster variations present in each cluster corresponding to each class are then preserved using symbolic interval data. Thus referenced logo images are represented in the form of interval data. A sample logo image is then classified using suitable symbolic classifier. For experimentation purpose, relatively large amount of color logo images are created consisting of 5044 logo images. The classification results are validated with the help of accuracy, precision, recall, f-measure, and time. To check the efficacy of the proposed model, the comparative analyses are given against the other models. The results show that the proposed model outperforms the other models with respect to time and f-measure.

**Keywords:** Appearance based features, Clustering, Symbolic Representation, Symbolic Classification, Logo image classification,

## 1 Introduction

With the rapid development of multimedia information technology, the amount of image data available on the internet is very huge and it is increasing exponentially. Handling of such a huge quantity of image data has become a more challenging and at the same time it is an interesting research problem. Nowadays, to handle such image data, there are lot many tools available on the internet such as ArcGIS, Google, Yahoo, Bing etc. Currently, those tools perform classification, detection, and retrieval of images based on their characteristics. In this work, we consider logos which come under image category for the purpose of classification. A logo is a symbol which symbolizes the functionalities of an organization.

Once a logo is designed for any organization, it needs to be tested for its originality and uniqueness. If not, many intruders can design logos which look very similar to the existing logos and may change the goodness of the respective organization. To avoid such trade infringement or duplication, a system to test a newly designed logo for its originality is required. To test for the originality, the system has to verify the newly designed logo by comparing with the existing logos. Since the number of logos available for comparison is very large, either a quick approach for comparison or any other alternative need to be investigated. One such alternative is to identify the class of logos to which the newly designed logo belongs and then verifying it by comparing against only those logos of the corresponding class. Thus, the process of classification reduces the search space of a logo verification system to a greater extent. With this motivation, we address a problem related to classification of logos based on their appearance.

### 1.1 Related Works

In literature, we can find couple of works carried out on logo classification. Especially, these works are mainly relied on black and white logo images.

In [1], an attempt towards classifying the logos of the University of Maryland (UMD) logo database is made. Here, the logo images are classified as either degraded logo images or non-degraded logo images. In [2], a logo classification system is proposed for classifying the logo images captured through mobile phone cameras with a limited set of images. In [3], a comparative analysis of invariant schemes for logo classification is presented. From the literature, it can be observed that, in most of the works, the classification of logos has been done only on logos present in the document images. Also, there is no work available for classification of color logo images. Keeping this in mind, we thought of classifying the color logos.

In this paper, an approach based on symbolic representation of features for classification of logo images is addressed. The logo images which represent the functionalities of organizations are categorized into three classes viz, images fully text, images with fully symbols, and images containing both texts and symbols. Some of these color logo images are illustrated in figure 1. Among three classes, there exist samples with large intra class variations as illustrated in figure 1. In figure 1, the both category logo images have several intra class variations like: the shape of text, the color of text, and the texture of symbols. In the second category of logo images, the internal variations are: the color of the text and the shape of text. In the last category i.e, symbols, the variations are with respect to the shapes of symbols measured at different orientations. The intra-class variations may lead to the mis-classification of a logo image as a member of given three classes. To avoid such mis-classification, methods which can take care of preserving the intra class variations present in the logo images is needed. One such method is symbolic data analysis. In symbolic data analysis, the interval representation of data can take care of intra class variations of features which help in classification [4], clustering [5], and regression [6] of data.

Due to the presence of very large logo image samples in every class of the dataset used, it is better to group the similar looking logo images in every class. This results with the class containing logos with lesser variations within the class. It further helps in representing them in symbolic interval form.

The paper mainly concentrates on only classification. In the literature, there exist some symbolic classifiers [7, 8]. But, the limitation of these classifiers is present at classification stage. As these classifiers make use of interval representation of both the reference samples and query samples while classification. But in our case, the reference logo images are represented as interval data and the query images are represented as a conventional crisp type data [9]. So to handle such data, a suitable symbolic classifier [9] which can take care of the above said scenario is used for classification.

Finally, the proposed system is compared with the other models viz., a model which make use of only clustering and a model which neither uses clustering nor uses symbolic representation. Our system has outperformed with the former and latter models in terms of validity measures and time respectively.

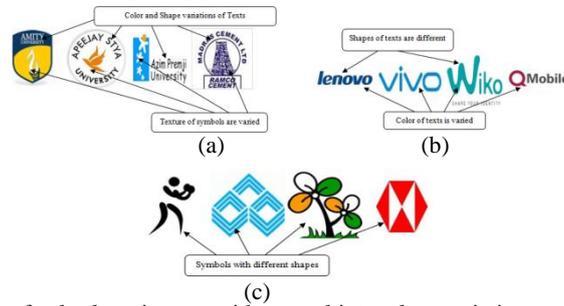
(a)　　　　　　　　　　　(b)

(c)
**Fig. 1.** Illustration of color logo images with several intra class variations exists with respect to (a) Both class, (b) only text class and (c) only symbol class.

The rest of the paper is organized as follows. In section 2, the details of the proposed logo classification system are explained. The experimentation setup and the detailed results with reasoning are presented in section 3. Further, comparisons to the proposed model against the conventional models are given. Finally, the conclusion remarks are given in section 4.

## 2　Proposed Model

Different steps involved in the proposed logo classification model are shown in figure 2. Our model is based on classifying a color logo as either a logo with only text or a logo with only symbols or a logo with both text and symbol. The different stages of the proposed model are explained in the following subsections.

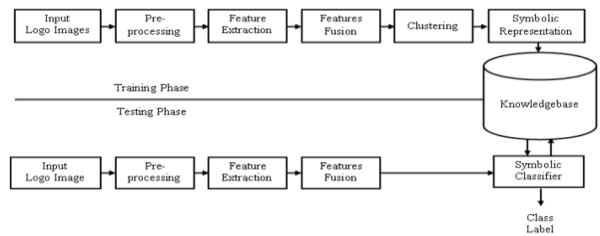
**Fig. 2.** Architecture of the proposed model

### 2.1 Pre-Processing

In this stage, we recommend two different pre-processing tasks, namely, image resizing and gray scale conversion. Initially, we resize all the logo images of dimension M x N into m x n to maintain the uniformity in the dimensions of the logo images. Where the former dimension relates with original logo images and latter dimension relates with resized logo images. Then, we convert the RGB logo images into gray scale images. The conversion helps in extracting the texture and shape features from gray scale images [10].

### 2.2 Feature Extraction and Fusion

In this work, we recommend three different appearance based (global) features namely, color, texture and shape features for extraction from an input color logo image. These features are recommended, as these features are invariant to geometrical transformations [10]. Color, texture, and shape features are extracted for all color logo images as explained in [11].

For color feature extraction, the resized RGB logo image is divided into eight coarse partitions or blocks and determines the mean and percentage of individual block with respect to each color component (red / blue / green). For texture and shape feature extraction, the gray scale logo images are used because these two features do not depend on the information of color properties of an image [10]. Initially for texture feature extraction, an image is processed using Steerable Gaussian filter decomposition with four different orientations ($0^0$, $-45^0$, $+45^0$, $90^0$), then the mean and standard deviation at each decomposition is computed. For shape features, Zernike moments shape descriptor is used in two different orientations ($0^0$ and $90^0$) [11]. Further, these features are fused together to discriminate the logo images.

### 2.3 Clustering

In this step, the logo images which look similar are grouped together with respect to each class. For clustering the similar logo images, the partitional clustering approach is adopted. The partitional clustering approach is very simple as it makes use of feature matrix for clustering instead of proximity matrix as in Hierarchical clustering [12]. Hence, in concern with the pre-processing efficiency of the proposed model, partitional clustering is chosen over hierarchical clustering.

The K-means clustering algorithm [12] is used to cluster the similar logo images. The value of K is varied to group the similar looking color logos within each class. The clustering results play a significant role in our proposed model in deciding the goodness of the proposed model.

### 2.4 Symbolic Logo Representation

In this step, the clustered logo images within each class are further represented in the form interval valued data, a representation which preserves the intra-class (cluster) variations [9]. The details of interval representation are given below.

Consider a sample $X_i = \{x^1, x^2, ..., x^d\}$ ($X_i \in R^d$), belongs to $i^{th}$ class containing $d$ features. Let there be totally $N$ number of samples from $m$ number of classes. If clustering is applied on the samples belong to $i^{th}$ class. The number of clusters obtained from each class is $k$. Then, the total number of samples present in $j^{th}$ cluster belongs to class $i$ be $n_j^i$ *(j=1,2,...,k and i=1,2,...,m)*. To preserve the intra class-variations present in each cluster, the mean-standard deviation interval representation is recommended. As it preserves the internal variations present within the samples of each clusters [9]. The mean and standard deviation computed for the clustered samples is given by (1) and (2).

$$\mu_{j_i}^l = \frac{1}{n_j^i} \sum_{h=1}^{n_j^i} x_h^l \tag{1}$$

$$\sigma_{j_i}^l = \sqrt{\frac{1}{(n_j^i - 1)} \sum_{h=1}^{n_j^i} (x_h^l - \mu_{j_i}^l)^2} \tag{2}$$

Where, $\mu_{j_i}^l$ and $\sigma_{j_i}^l$ are the mean and standard deviation value of $l^{th}$ feature belongs to $j^{th}$ cluster corresponding to class $i$ respectively.

Further, the mean and standard deviation is computed for all features belongs to $j^{th}$ cluster corresponding to $i^{th}$ class.

After computing the mean and standard deviation for each cluster belongs to a respective class. These two moments are joined together to form an interval cluster representative belongs to each class. The difference between mean and standard deviation represents lower limit of an interval and the sum of mean and standard deviation represents the upper limit of an interval. Finally, $k$ number of such cluster interval representatives are obtained from each class. Cluster representative is given by:

$$CR_j^i = \{[(\mu_{j_i}^1 - \sigma_{j_i}^1), (\mu_{j_i}^1 + \sigma_{j_i}^1)], [(\mu_{j_i}^2 - \sigma_{j_i}^2), (\mu_{j_i}^2 + \sigma_{j_i}^2)], ..., [(\mu_{j_i}^d - \sigma_{j_i}^d), (\mu_{j_i}^d + \sigma_{j_i}^d)]\}$$

$$CR_j^i = \{[f_1^-, f_1^+], [f_2^-, f_2^+], ..., [f_d^-, f_d^+]\}$$

$$where, f_l^- = \{(\mu_{j_i}^l - \sigma_{j_i}^l)\} \text{ and } f_l^+ = \{(\mu_{j_i}^l + \sigma_{j_i}^l)\}$$

Finally, we arrived at an interval feature matrix of dimension *(k\*m)xd*, considered as a reference matrix while classification.

### 2.5 Logo Classification

To test the effectiveness of the proposed classification system, suitable symbolic classifier is needed. Here, we make use of a symbolic classifier proposed in [9] for classification of logo images. Here, the reference logo images are represented in the form of interval data as explained in the earlier sections. Let us consider a test sample $S_q = \{s^1, s^2, ..., s^d\}$, contains $d$ number of features. The test sample $S_q$ needs to be classified as a member of any one of the three classes. Hence, the similarity is computed between a test sample and all reference samples. Here, for every test sample the similarity is computed at feature level. So, the similarity between a test crisp (single valued) feature and a reference interval feature can be computed as follows: The similarity value is 1, if the crisp value lies between the upper limit and lower limit of an interval feature, else 0. Similarly, the similarity between $S_q$ and all remaining samples are computed. If $S_q$ is said to be a member of any one of the three classes, then the value of acceptance count $AC_q^{j_i}$ is very high with respect to the reference sample (cluster representative) belongs to a particular class.

The acceptance count $AC_q^{j_i}$ for a test sample corresponding to $j^{th}$ cluster of $i^{th}$ class is given by:

$$AC_q^{j_i} = \sum_{l=1}^{d} Sim(S_q, CR_j^i) \qquad (3)$$

Where, $Sim(S_q, CR_j^i) = \begin{cases} 1 & if\ s^l \geq f_l^- \ and\ s^l \leq f_l^+ \\ 0 & otherwise \end{cases}$ and

$i=1,2,…,m;\ j=1,2,…,k;\ and\ l=1,2,…,d$

## 3   Experimentation

### 3.1   Dataset

For experimentation, we have created our own dataset named "UoMLogo database" consisting of 5044 color logo images. The "UoMLogo Database" mainly consists of color logo images of different universities, brands, sports, banks, insurance, cars, and industries etc. which are collected from the internet. This dataset mainly categorized into three classes, BOTH logo image (a combination of TEXT and SYMBOL), TEXT logo image, SYMBOL image. Within class, there exist ten different subclasses. Figure 3 shows the sample images of the UoMLogo Dataset. The complete details of the dataset found in [11].

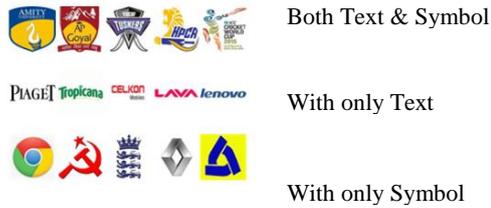

Both Text & Symbol

With only Text

With only Symbol

**Fig. 3.** Samples logo images of UoMLogo Dataset

### 3.2   Experimental Setup

In preprocessing step, for the sake of simplicity and uniformity in extracting the features from a color logo image, we have resized every image into 200 x 200 dimensions. To extract texture and shape features, the color logo images are converted into grayscale images, as these two features are independent of color features. In feature extraction stage, three different features mainly: color, texture and shape features are extracted. These features are extracted from logo images as discussed in the section 2.2. From feature extraction, we arrived at 48, 8, and 4 color, texture, and shape features respectively. Further, these features are normalized and fused to get 60 features, which represent a logo image sample.

After feature extraction and fusion, these features of logo images are represented in the form of feature matrix. Where the rows and the columns of a matrix represent the samples and features respectively. Further, the samples of a feature matrix are divided into training samples and testing samples. Training samples are used for clustering and symbolic representation. Testing samples are used for testing the classification system.

During symbolic representation, the training samples are clustered using K-means algorithm. The values of K varied from 2 to 10, depending on the formation of cluster samples. The upper bound of the clusters is limited to 10, because; the clustering algorithm fails to cluster the similar logo images beyond 10. Further, the clustered samples are represented in the form of interval data as explained in section 2.5. The total number of logo images present in our database is 5044 with 3 different classes. So, the total number of samples present in a reference matrix after symbolic representation is 6 (2x3=6; 2: number of cluster interval representative; 3: No. of Classes), 9, 12, 15, 18, 21, 24, 27, and 30 samples for clusters varied from 2 to 10 respectively. For classification, a symbolic classifier is adopted.

In our proposed classification system, the dataset is divided randomly into training and testing. Seven sets of experiments have been conducted under varying number of training set images as 20%, 30%, 40%, 50%, 60%, 70% and 80%. At each training stage, the logo images are represented in the form of interval data with respect to the varied number of clusters from 2 to 10. While at testing stage, the system uses remaining 80%, 70%, 60%, 50%, 40%, 30%, and 20% of logo images respectively for classifying them as any one of the three classes. The experimentation in testing is repeated for 20 different trials. During testing, the classification results are presented by the confusion matrix. The performance of the classification system is evaluated using classification accuracy, precision, recall, and F-Measure computed from the confusion matrix [13].

### 3.3 Experimental Results

The performance of the proposed classification system is evaluated not only based on classification accuracy, precision, recall, and F-Measure computed from the confusion matrix but also it is done with respect to time.

Let us consider a confusion matrix $CM_{ij}$, generated during classification of color logo images at some testing stage. From this confusion matrix, the accuracy, the precision, the recall, and the F-Measure are all computed to measure the efficacy of the proposed logo image classification system. The overall accuracy of a system is given by:

$$Accuracy = \frac{No.of\ Correctly\ classified\ Samples}{Total\ number\ of\ Samples} * 100 \qquad (4)$$

The precision and recall can be computed in two ways. Initially, they are computed with respect to each class and later with respect to overall classification system. The class wise precision and class wise recall is computed from the confusion matrix are given in equations (5) and (6) respectively.

$$P_i = \frac{No.\,of\ Correctly\ classified\ Samples}{No\ of\ Samples\ classified\ as\ a\ member\ of\ a\ class} * 100 \quad (5)$$

$$R_i = \frac{No.\,of\ Correctly\ classified\ Samples}{Expected\ number\ of\ Samples\ to\ be\ classified\ as\ a\ member\ of\ a\ class} * 100 \quad (6)$$

Where, i=1,2,…,m; m=No. of Classes

The system precision and system recall computed from the class wise precision and class wise recall is given by:

$$Precision = \frac{\sum_{i=1}^{m} P_i}{m} \quad (7)$$

$$Recall = \frac{\sum_{i=1}^{m} R_i}{m} \quad (8)$$

The F-measure computed from the precision and recall is given by:

$$F - Measure = \frac{2*Precision*Recall}{Precision+Recall} * 100 \quad (9)$$

The average time is computed at a particular testing stage while classifying a test sample as a member of any one of the three classes. It is done using a MATLAB built in command *tic – toc*.

The classification results are thus obtained for different training and testing percentage of samples under varied clusters from 2 to 10. These results are measured in terms of accuracy (minimum, maximum, and average), precision (minimum, maximum, and average), recall (minimum, maximum, and average) and F-Measure (minimum, maximum, and average). The minimum, maximum, and average of respective results are obtained due to the 20 trials of experiments performed on training samples. Here, precision and recall are computed from the results obtained from the class wise precision and class wise recall respectively.

The results obtained from cluster 2 to 10 under varied training and testing samples are consolidated and tabulated in table 1. These results are judged based on the best average F-measure obtained under respective training and testing percentage among all clusters (varied from 2 to 10). This has been clearly shown in figure 4.

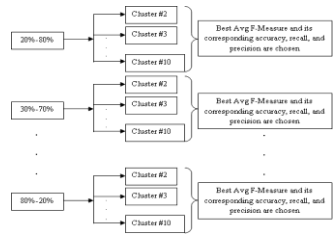

**Fig. 4.** Selection procedure followed in choosing the best results obtained from all clusters under varied training and testing percentage of samples

Table 1. Best results obtained from all clusters under varied training and testing percentage of samples

| Train-Test % | Accuracy | | | Precision | | | Recall | | | F-Measure | | | Cluster # |
|---|---|---|---|---|---|---|---|---|---|---|---|---|---|
| | Min | Max | Avg | Min | Max | Avg | Min | Max | Avg | Min | Max | Avg | |
| 20-80 | 60.20 | 74.21 | 68.56 | 52.77 | 81.95 | 65.35 | 53.85 | 58.48 | 56.07 | 54.78 | 65.39 | 60.14 | 10 |
| 30-70 | 60.90 | 74.07 | 68.28 | 55.55 | 77.19 | 64.06 | 55.07 | 58.91 | 56.92 | 56.33 | 64.93 | 60.13 | 9 |
| 40-60 | 61.92 | 74.38 | 70.47 | 56.78 | 80.08 | 65.74 | 54.67 | 59.16 | 56.65 | 56.21 | 65.77 | 60.71 | 5 |
| 50-50 | 60.97 | 73.70 | 69.81 | 59.05 | 71.22 | 65.06 | 55.79 | 60.28 | 57.45 | 57.88 | 63.14 | 60.95 | 10 |
| 60-40 | 61.71 | 73.61 | 70.75 | 58.08 | 69.76 | 65.11 | 56.16 | 59.77 | 57.52 | 57.10 | 63.45 | 61.03 | 5 |
| **70-30** | **66.27** | **73.94** | **71.73** | **58.70** | **72.94** | **66.89** | **55.38** | **59.29** | **57.27** | **58.09** | **64.20** | **61.63** | **4** |
| 80-20 | 65.77 | 73.31 | 70.23 | 59.27 | 72.21 | 64.55 | 56.41 | 61.22 | 58.34 | 58.70 | 63.91 | 61.24 | 8 |
| **Best** | **66.27** | **73.94** | **71.73** | **58.70** | **72.94** | **66.89** | **55.38** | **59.29** | **57.27** | **58.09** | **64.20** | **61.63** | **4** |

From the above table, it is very clear that the best classification results are obtained only if the samples within each class are clustered into 4 groups. This shows that our model is very robust in choosing the number of clusters for clustering samples within each class. The last row in the above table reveals the best results obtained for respective cluster and at for respective training and testing percentage of samples.

### 3.4 Comparative Analyses

The proposed symbolic logo image classification model is compared against the two different models in classifying the same color logo image database. As we know our model makes use of clustering before representing the samples in symbolic representation. We thought of comparing our model against the other models: a) a model which never preserves intra-class variations and never groups the similar logo images, and b) a model which makes use of clustering for grouping similar logo images within each class but does not preserve intra-class variations (i.e., this model never make use of symbolic representation, it only uses with conventional cluster mean representation). This comparison helps us to test robustness of the proposed model in terms of F-measure and time.

The experimental setup for the other two models is followed as given below:

For Conventional Model (Model-1):

Totally 60 features are considered for representation. K-NN classifier (K=1) is used for classification. The training and testing percentage of samples are divided and varied from 20% to 80% (in steps of 10% at a time). The experiments are repeated for 20 trials and results are noted based on the minimum, maximum, and average values obtained from 20 trials.

For Conventional + Clustering (Co+Cl) Model (Model-2):

Totally 60 features are considered for representation. Then K-means partitional clustering algorithm is applied to group similar logo images within each class (K: varied from 2 to 10). K-NN classifier (K=1) is used for classification. The training and testing percentage of samples are divided and varied from 20% to 80% (in steps of 10% at a time). The experiments are repeated for 20 trials and results are noted based on the minimum, maximum, and average values obtained from 20 trials.

The classification results for the above said models are validated based on the confusion matrix obtained during the classification. The same validity measures used in our proposed model (accuracy, precision, recall, F-measure, and time) are used for validating these two models. With respect to model-1, the best results are obtained for 80-20 percent of training and testing samples and are shown in table 2. With respect to model-2, the best results are obtained when the similar logo samples are grouped into two clusters. The results are tabulated in table 3. Similarly, the results of the proposed model are tabulated in table 4.

**Table 2.** Best results obtained from different training and testing percentage of samples based on Avg F-measure

|  | Min | Max | Avg | Train-Test % |
|---|---|---|---|---|
| **Accuracy** | 69.31 | 73.51 | 71.48 | 80-20 |
| **Precision** | 59.40 | 66.72 | 62.36 | |
| **Recall** | 58.46 | 65.12 | 60.94 | |
| **F-Measure** | 59.09 | 65.43 | 61.64 | |

**Table 3.** Best results obtained for Conventional + Clustering Classification model

|  | Min | Max | Avg | Cluster # | Train-Test % |
|---|---|---|---|---|---|
| **Accuracy** | 53.08 | 60.02 | 56.97 | 2 | 80-20 |
| **Precision** | 42.98 | 50.75 | 47.57 | | |
| **Recall** | 44.08 | 53.63 | 49.58 | | |
| **F-Measure** | 43.53 | 51.88 | 48.55 | | |

**Table 4.** Best results obtained for Symbolic + Clustering Classification model

|  | Min | Max | Avg | Cluster # | Train-Test % |
|---|---|---|---|---|---|
| **Accuracy** | 66.27 | 73.94 | 71.73 | 4 | 70-30 |
| **Precision** | 58.70 | 72.94 | 66.89 | | |
| **Recall** | 55.38 | 59.29 | 57.27 | | |
| **F-Measure** | 58.09 | 64.20 | 61.63 | | |

The results shown in the Tables 2, 3, and 4 are for the respective models in classifying the color logo images. From Tables 2 and 3, it is very clear that the model-1 is superior model compared to model-2 in terms of average F-measure and remaining other measures. But in terms of efficiency in classification, model-2 outperforms model-1 as shown in Fig. 5. With respect to model-2 and proposed, it is clearly observed from Tables 3 and 4 that the proposed model is far superior than the model-2 in terms of average F-measure (and other remaining measures), and also in terms of efficiency, our model is very stable model irrespective of training and testing percentage of samples. Similarly, if we compare model-1 with our own model, our model is somehow equivalent to model-1 with an epsilon difference in average F-measure. But, if we consider with respect to efficiency in classification, our model outperforms model-1 in terms of stability and efficiency in classification. Hence, the proposed model suits better for classifying the huge color logo image database.

For better visualization on classification of color images, the confusion matrices obtained for the best results shown in Tables 3 and 4 are given in Tables 5 and 6, respectively; and also, the misclassifications occurred while classifying the logo images for model-2 and proposed model are given in Fig. 6a, b, respectively.

Table 5. Confusion matrix obtained for model-2
(Conventional + Clustering method for Cluster #2 (80%-20%))

|              | Both | Text | Symbol |
|---|---|---|---|
| Both (634)   | **399** | 137  | 98     |
| Text (249)   | 87   | **118** | 44     |
| Symbol (125) | 63   | 23   | **39** |

Table 6. Confusion matrix obtained for the proposed model
(Symbolic + Clustering method for Cluster #4 (70%-30%))

|              | Both | Text | Symbol |
|---|---|---|---|
| Both (951)   | **818** | 86   | 47     |
| Text (419)   | 154  | **194** | 25     |
| Symbol (188) | 93   | 24   | **71** |

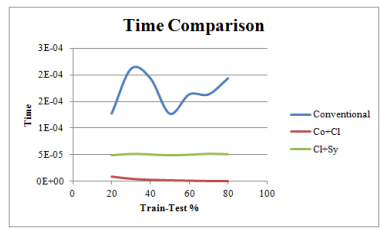

**Fig. 5.** Comparison of Time utilization of the proposed model vs conventional model vs conventional + clustering model in classifying the logo images.

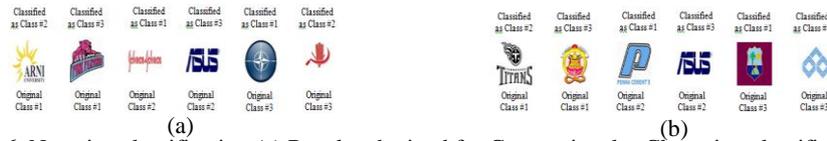

(a)                         (b)
**Fig. 6.** Negative classification (a) Results obtained for Conventional + Clustering classification model; (b) Results obtained for Symbolic + Clustering classification model

From the above discussion, it is very very clear that the proposed logo image classification system is very stable and efficient compared to other models in classifying the color logo images.

## 4 Conclusion

In this paper an approach based on symbolic interval representation in classifying the color logo images into the pre-defined three classes is proposed. In classifying a logo image, the global characteristics of logo images are extracted. Then, the partitional clustering algorithm is adopted for clustering the similar logo images within each class. Later, the symbolic interval representation is given for clusters belong to the corresponding classes. Further, a symbolic classifier is used for logo image classification. The effectiveness of the proposed classification system is validated through well known measures like accuracy, precision, recall, F-Measure and also with respect to time. Finally, the paper concludes with an understanding that the better classification results are obtained only for the symbolic interval representation model compared to other models.

**Acknowledgements.** The author N Vinay Kumar would like to thank Dept. of Science & Technology, India, for the financial support through INSPIRE Fellowship.